\documentclass[11pt,a4paper]{article}
\usepackage[hyperref]{acl2020}
\usepackage{times}
\usepackage{latexsym}

\usepackage{microtype}
\usepackage{graphicx}
\usepackage{booktabs}
\usepackage{float}
\usepackage{subfigure}
\usepackage{array}
\newcolumntype{L}[1]{>{\raggedright\let\newline\\\arraybackslash\hspace{0pt}}m{#1}}
\newcolumntype{C}[1]{>{\centering\let\newline\\\arraybackslash\hspace{0pt}}m{#1}}
\newcolumntype{R}[1]{>{\raggedleft\let\newline\\\arraybackslash\hspace{0pt}}m{#1}}

\aclfinalcopy

\title{ The NiuTrans System for WNGT 2020 Efficiency Task }

\author{Chi Hu$^{\dagger}$ \,
  Bei Li$^{\dagger}$ \,
  Ye Lin$^{\dagger}$ \,
  Yinqiao Li$^{\dagger}$ \\
  {\bf Yanyang Li$^{\dagger}$ \,
  Chenglong Wang$^{\dagger}$ } \\
  {\bf Tong Xiao$^{\dagger}$$^{\ddagger}$\,
  Jingbo Zhu$^{\dagger}$$^{\ddagger}$ } \\
  $^{\dagger}$NLP Lab, Northeastern University, Shenyang, China \\
  $^{\ddagger}$NiuTrans Reasearch, Shenyang, China \\
  {\tt
        huchinlp@gmail.com, libei\_neu@outlook.com,
        }\\
    {\tt
        \{xiaotong,zhujingbo\}@mail.neu.edu.com,
        } \\
  }

\date{}

\begin{document}
\maketitle
\begin{abstract}
This paper describes the submissions of the NiuTrans Team to the WNGT 2020 Efficiency Shared Task. We focus on the efficient implementation of deep Transformer models \cite{wang-etal-2019-learning, li-etal-2019-niutrans} using NiuTensor\footnote{\url{https://github.com/NiuTrans/NiuTensor}}, a flexible toolkit for NLP tasks. We explored the combination of deep encoder and shallow decoder in Transformer models via model compression and knowledge distillation. The neural machine translation decoding also benefits from FP16 inference, attention caching, dynamic batching, and batch pruning. Our systems achieve promising results in both translation quality and efficiency, e.g., our fastest system can translate more than 40,000 tokens per second with an RTX 2080 Ti while maintaining 42.9 BLEU on \textit{newstest2018}. The code, models, and docker images are available at NiuTrans.NMT\footnote{\url{https://github.com/NiuTrans/NiuTrans.NMT}}.
\end{abstract}

\section{Introduction}
In recent years, the Transformer model and its variants \cite{Vaswani2017AttentionIA, Shaw2018SelfAttentionWR, So2019TheET, Wu2019PayLA, wang-etal-2019-learning} have established state-of-the-art results on machine translation (MT) tasks. However, achieving high performance requires an enormous amount of computations \cite{Strubell2019EnergyAP}, limiting the deployment of these models on devices with constrained hardware resources. 

The efficiency task aims at developing MT systems to achieve not only translation accuracy but also memory efficiency or translation speed across different devices. This competition constraints systems to translate 1 million English sentences within 2 hours. Our goal is to improve the quality of translations while maintaining enough speed. We participated in both CPUs and GPUs tracks in the shared task. 

Our system was built with NiuTensor, an open-source tensor toolkit written in C++ and CUDA based on dynamic computational graphs. NiuTensor is developed for facilitating NLP research and industrial deployment. The system is lightweight, high-quality, production-ready, and incorporated with the latest research ideas. 

We investigated with a different number of encoder/decoder layers to make trade-offs between translation performance and speed. We first trained several strong teacher models and then compressed teachers to compact student models via knowledge distillation \cite{Hinton2015DistillingTK, kim-rush-2016-sequence}. We find that using a deep encoder (up to 35 layers) and a shallow decoder (1 layer) gives reasonable improvements in speed while maintaining high translation quality. We also optimized the Transformer model decoding in engineering, such as caching the decoder's attention results and using low precision data type.

We present teacher models and training details in Section \ref{sec:teacher models}, then in Section \ref{sec:student models} we describe how to obtain lightweight student models for efficient decoding. Optimizations for the decoding across different devices are discussed in Section \ref{sec:optimizations}. We show the details of our submissions and the results in Section \ref{sec:submissions}. Section \ref{sec:conclusion} 
summarizes this paper and describes future work.

\section{Deep Transformer Teachers}
\label{sec:teacher models}

\subsection{Deep Transformer Architectures}
Recent years have witnessed the success of transformer-based models in MT tasks. Many works \cite{Dehghani2019UniversalT, Zhang2019ImprovingDT, Li2020NeuralMT} focus on designing new attention mechanisms and Transformer architectures. \citet{Shaw2018SelfAttentionWR} extended the self-attention to consider the relative position representations or distances between words. \citet{Wu2019PayLA} replaced the self-attention components with lightweight and dynamic convolutions. Deep Transformer models also attracted a lot of attention. \citet{Wang2018MultilayerRF} proposed a multi-layer representation fusion approach to learn a better representation from the stack. \citet{wang-etal-2019-learning} analyzed the high risk of gradient vanishing or exploring in the standard Transformer, which place the layer normalization \cite{Ba2016LayerN} after the attention and feed-forward components. They showed that a deep Transformer model can surpass the big one by proper use of layer normalization and dynamic combinations of different layers. In their method, the input of layer $l + 1$ is defined by:
\begin{equation}x_{l+1}=\mathcal{G}\left(y_{0}, \dots, y_{l}\right)\end{equation}
\begin{equation}\mathcal{G}\left(y_{0}, \ldots, y_{l}\right)=\sum_{k=0}^{l} W_{k}^{(l+1)} \mathrm{L} \mathrm{N}\left(y_{k}\right)\end{equation}
where $y_{l}$ is the output of the $l_th$ layer and $W$ is the weights of different layers.

We employed the dynamic linear combination of layers Transformer architecture incorporated with relative position representations as our teacher network, call it Transformer-DLCL-RPR. 
%

\subsection{Training Details}
We followed the constrained condition of the WMT 2019 English-German news translation task and used the same data filtering method as  \cite{li-etal-2019-niutrans}. We also normalized punctuation and tokenized all sentences with the Moses tokenizer \cite{Koehn2007MosesOS}. The training set contains about 10M sentences pairs after processed. In our systems, the data was tokenized, and jointly byte pair encoded \cite{Sennrich2016NeuralMT} with 32K merge operations using a shared vocabulary. After decoding, we removed the BPE separators and de-tokenize all tokens.

We trained four teacher models using \textit{newstest2018} as the development set with fairseq \cite{Ott2019fairseqAF}. Table \ref{table:teachers} shows the results of all teacher models and their ensemble, where we report SacreBLEU \cite{post-2018-call} and the model size. 
The difference between teachers is the number of encoder layers and whether they contain a dynamic linear combination of layers. All teachers have 6 decoder layers, 512 hidden dimensions, and 8 attention heads. We shared the source-side and target-side embeddings with the decoder output weights. The maximum relative length was 8, and the maximum position for both source and target was 1024. We used the Adam optimizer \cite{Kingma2015AdamAM} with $\beta_1 = 0.9$, $\beta_2 = 0.997$ and $\epsilon = 10^{-8}$ as well as gradient accumulation due to the high GPU memory footprint. Each model was trained on 8 RTX 2080Ti GPUs for up to 21 epochs. We batched sentence pairs by approximate length and limited input/output tokens per batch to 2048/GPU. Following the method of \cite{wang-etal-2019-learning}, we accumulated every two steps for a better batching. This resulted in approximately 56000 tokens per training batch. The learning rate was decayed based on the inverse square root of the update number after 16000 warm-up steps, and the maximum learning rate was 0.002. Furthermore, we averaged the last five checkpoints in the training process for all models. 

\begin{table}
\centering
\begin{tabular}{lrr}
\hline \textbf{Model} & \textbf{Param.} & \textbf{BLEU} \\ \hline
Transformer-35-6 &  152M & 43.3 \\
Transformer-35-6+DLCL & 152M & 43.7 \\
Transformer-40-6 & 168M & 44.5 \\
Transformer-40-6+DLCL & 168M & 43.9 \\
Ensemble & 640M & 45.5 \\
\hline
\end{tabular}
\caption{\label{table:teachers} Results on \textit{newstest18} - Teacher Models. 35-6 means that the model contains 35 encoder layers and 6 decoder layers. }
\end{table}

As shown in Table \ref{table:teachers}, the best single teacher model achieves 44.5 BLEU (beam size 4) on \textit{newstest2018}. Then we obtained an improvement of 1 BLEU via a simple ensemble strategy used in \cite{li-etal-2019-niutrans}.

\section{Lightweight Student Models}
\label{sec:student models}
After the training of deep Transformer teachers, we compressed the knowledge in an ensemble into a single model through knowledge distillation \cite{Hinton2015DistillingTK, kim-rush-2016-sequence}. Then we analyzed the decoding time of each part in the deep Transformer. We further pruned the encoder and decoder layers to improve the decoding efficiency.

\subsection{Knowledge Distillation}
Knowledge distillation approaches \cite{Hinton2015DistillingTK, kim-rush-2016-sequence} have proven successful in reducing the size of neural networks. They learn a smaller student model to mimic the original teacher network by minimizing the loss between the student and teacher output. We applied the sequence-level knowledge distillation on the teacher ensemble described in Section \ref{sec:teacher models}. We used the ensemble to generate multiple translations of the raw English sentences. In particular, we collected the 4-best list for each sentence against the original target to create the synthetic training data.
Our base student model consists of 35 encoder layers and six decoder layers (call it 35-6) with nearly 150M parameters. It achieves 44.6 BLEU on the test set.

\begin{figure}
\begin{center}
\includegraphics[height=14em]{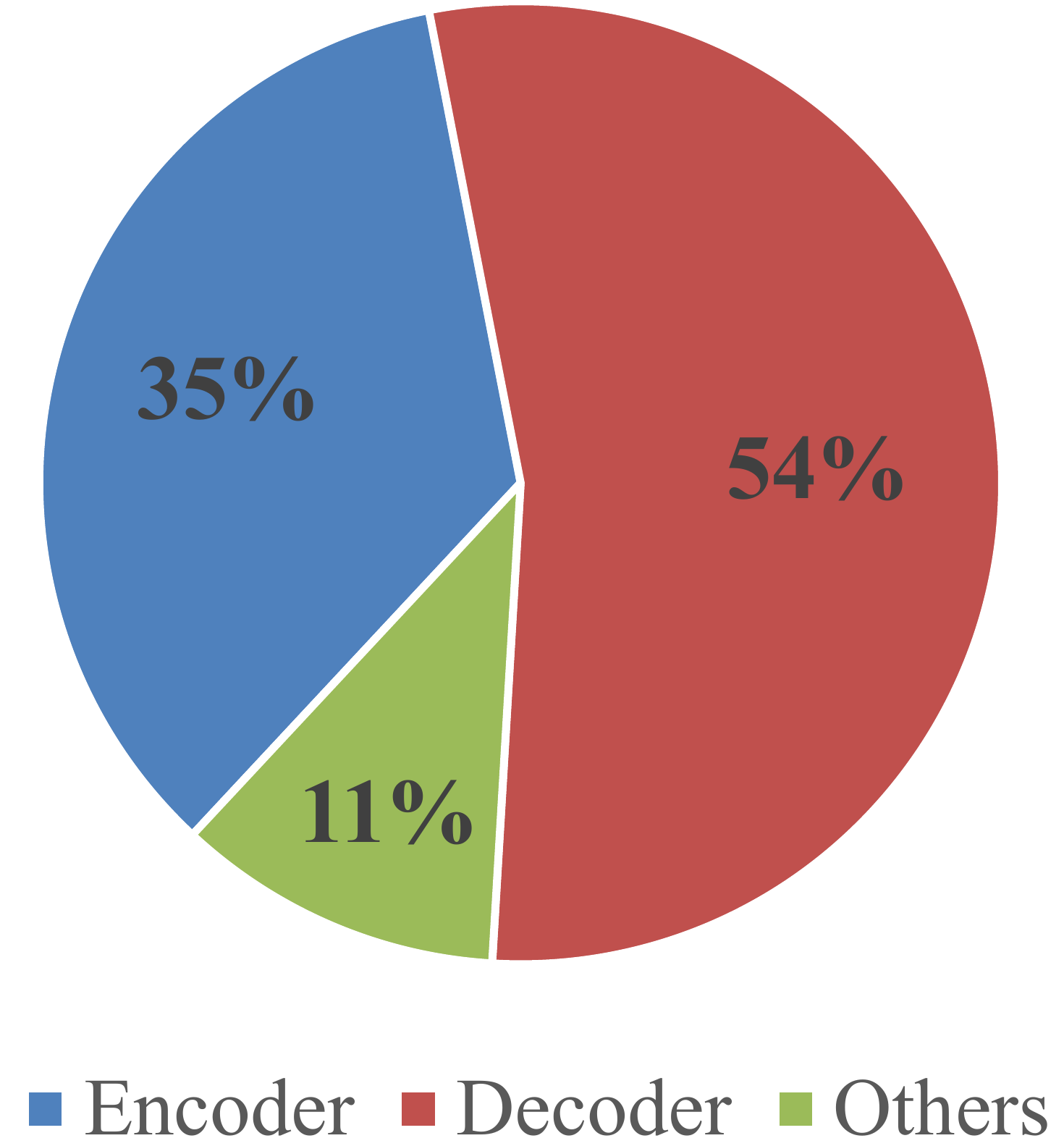}
\end{center}
\caption{Profiling of the throughput during inference on \textit{newstest2018} using a 35-6 model. }
\label{figure:decoding-time}
\end{figure}

\subsection{Fast Student Models}
Although the deep model can obtain high-quality translations, its speed is not satisfactory. For example, it costs 6.7 seconds to translate 2998 sentences on a 2080Ti GPU using a 35-6 model with the greedy search.
Statistics show that the most time-consuming part of the decoding process is the decoder, as presented in Figure \ref{figure:decoding-time}, so the most efficient optimization is to use a lightweight decoder. To make a comparison, we kept the 35 encoder layers and reduced the decoder layer to 1. In practice, we copied the bottom layers' parameters from big models to small models for initialization. Then we trained the small models as usual. Similar to \cite{wang-etal-2019-learning}, the encoder has a more significant influence on the translation quality than the decoder. Reducing the number of decoder layers brings us a speedup of more than 30\% with a slight loss of 0.3 BLEU. 

We further compressed the model by shrinking the encoder. Unless otherwise stated, the following student models have only one decoder layer. We copied the bottom layer parameters from big models to initialize small models to stabilize the training. We trained two small models with an 18-layer encoder and a 9-layer encoder, respectively. Table \ref{table:students} shows the comparison of different teachers and students.
Compared with the 35-1 model, cutting off half of the encoder layer reduces the parameters by nearly half and gives a speedup of 20\% with a decrease of 0.2 BLEU. The 9-1 model is the fastest model we run on the GPU. It can translate \textit{newstest2018} within 3 seconds on a 2080Ti GPU and obtain 42.9 BLEU. 

All models mentioned above can translate 1 million sentences on the GPU in 2 hours. However, using a CPU to achieve this goal is not easy, so we need smaller models. We set the 9-1 model size to 256 for the CPU version, namely 9-1-tiny, which has only half the 9-1 model parameters. This model achieves 37.2 BLEU on \textit{newstest2018} and reduces 90\% parameters compared to the 35-6 model. 

\begin{table}[t]
\begin{center}
\begin{tabular}{lrrr}
\hline \textbf{Model} & \textbf{Param.} & \textbf{Speedup} & \textbf{BLEU} \\ \hline
\textbf{Teacher-40-6} &  168M & 1x & 44.5 \\ \hline
Student-35-6 &  152M & 1.1x & 44.6 \\
Student-35-1 & 131M & 1.6x &  44.3 \\
Student-18-1 & 77M & 2.0x & 43.4 \\
Student-9-1 & 49M & 2.4x & 42.9 \\
Student-tiny & 25M & 2.9x & 37.2 \\
\hline
\end{tabular}
\end{center}
\caption{\label{table:students} Results on \textit{newstest18}. The students were trained by sequence-level knowledge distillation. The tiny setting keeps the 9-1 model's configurations except for using a model size of 256. We report the translation speed  on a single 2080Ti. }

\end{table}

\section{Optimizations for Decoding}
\label{sec:optimizations}

\subsection{General Optimizations}
First, we discuss some device-independent optimization methods. \\
\textbf{Caching} \quad
We can cache the output of the top layer of the encoder and each step of the decoder since we use an autoregressive model. More specifically, we cache the linear transformations for keys and values before the self-attention and cross-attention layers. \\
\noindent \textbf{Faster Beam Search} \quad
Beam search is a common approach in sequence decoding. The standard beam search strategy generates the target sequence in a self-regression manner and keeps a fixed amount of active candidates during decoding. We adopt a basic strategy to accelerate beam search: the search ends when any candidate predicts the EOS symbol, and there are no candidates with higher scores. This strategy brings us up to a 20\% speedup on the WMT test set. Other threshold-based pruning strategies \cite{Freitag2017BeamSS} are not appropriate due to the complex hyper-parameters.  \\
\textbf{Batch Pruning} \quad
The length of target sequences may vary for different sentences in a batch, which makes the computation inefficient. We prune the finished hypotheses in a batch during decoding but only gain little accelerations on CPUs.

\begin{figure*} 
  \centering 
  \subfigure[Operations before optimizing.]{ 
    \label{figure:operations} 
    \includegraphics[height=13em]{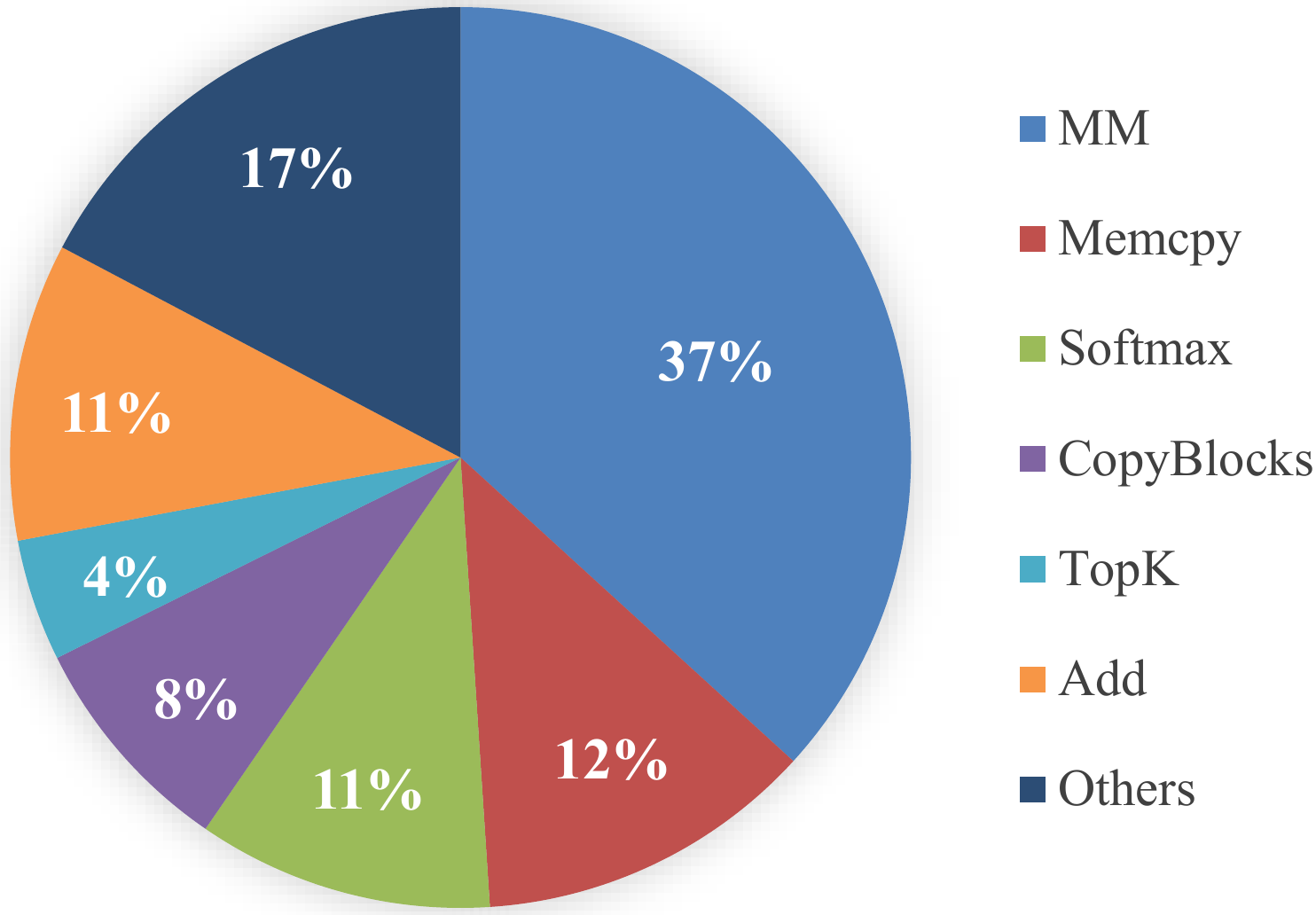}} 
  \subfigure[Operations after optimizing]{ 
    \label{figure:optimized_operations} 
    \includegraphics[height=13em]{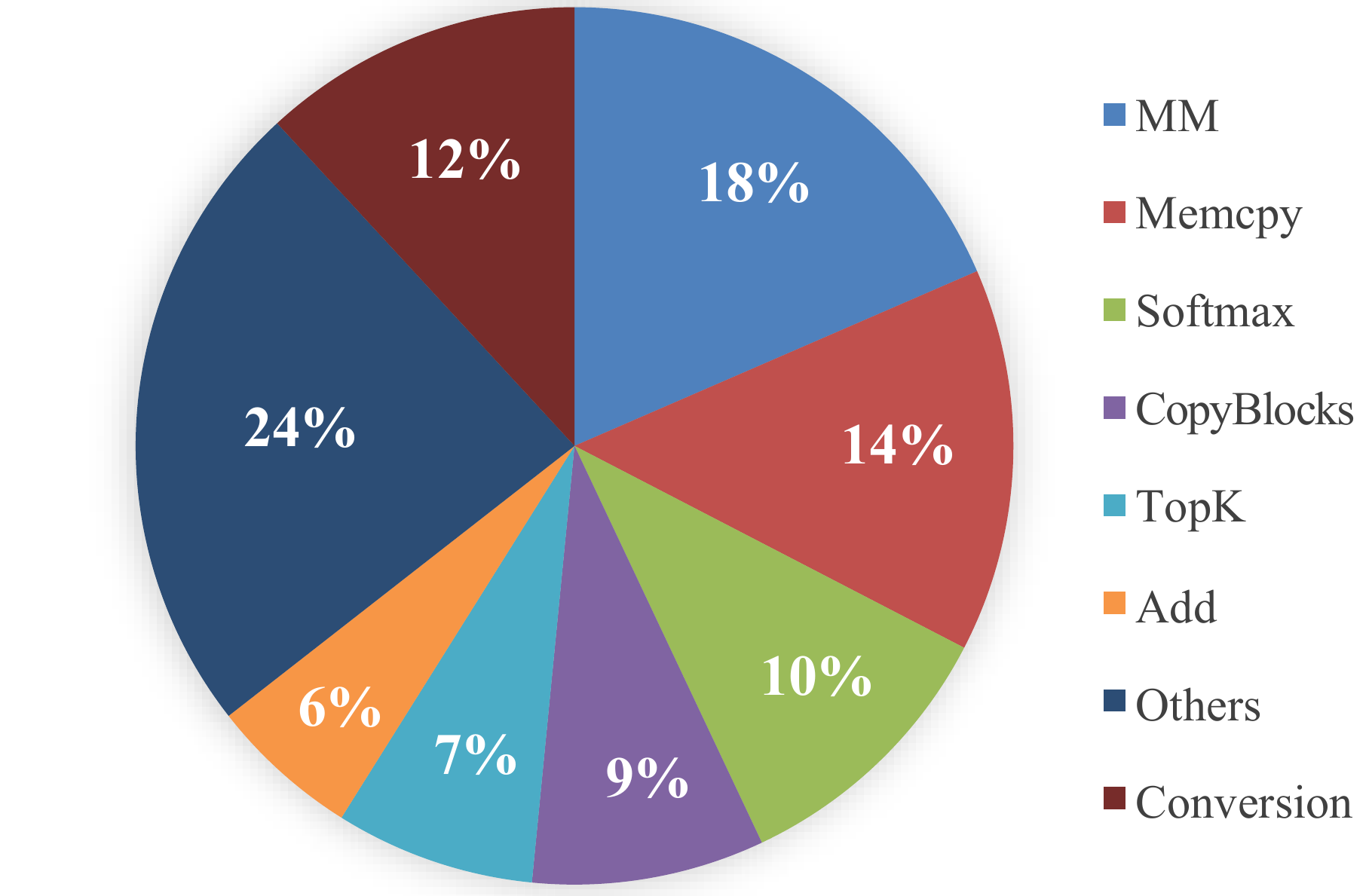}} 
  \caption{Profiling results of all operations during inference before or after optimizing on \textit{newstest2018} using a 9-1 model on a 2080Ti. We performed decoding for ten times to get more convincing results. Before optimizing, the decoding time is 76.9 seconds.  The combination of different optimizations reduces the time to 24.9 seconds. MM is matrix multiplication, and CopyBlocks is used in the tensor copy.} 
  \label{fig:subfig} 
\end{figure*}
\subsection{Optimizing for GPUs}
For the GPU-based decoding, we mainly explored dynamic batching, FP16 inference, and profiling. \\
\textbf{Dynamic Batching} \quad
Unlike the CPU version, the easiest way to reduce the translation time on GPUs is to increase the batch size within a specific range. We implemented a dynamic batching scheme that maximizes the number of sentences in the batch while limiting the number of tokens. This strategy significantly accelerates decoding compared to using a fixed batch size when the sequence length is short. \\
\textbf{FP16 Inference} \quad
Since the Tesla T4 GPU supports calculations under FP16, our systems execute almost all operations in 16-bit floating-point. All model parameters are stored in FP16, which reduces the model size on disk by half. We tried to run all operations at a 16-bit floating-point. However, in our test, some particular inputs will cause numerical instability, such as large batch size or sequence length. To escape overflow, we convert the data type around some potentially problematic operations, i.e., all operations related to $reduce\_sum$.

\subsection{Optimizing for CPUs}
As mentioned above, the goal we set for the CPU version is to translate 1 million sentences in 2 hours. We used the same settings as the 9-1 model except that the model size is 256 and therefore sacrifice about 6 BLEU on the WMT test set. We employed two methods to speed up the decoding on CPUs.\\
\textbf{Using of MKL} \quad
To make the full use of the Intel architecture and to extract the maximum performance, the NiuTensor framework is optimized using the Intel Math Kernel Library for basic operators. We can take advantage of this convenience with only minor changes to the configuration. \\
\textbf{Decoding in Parallel} \quad
The target machine in this task has 96 logical processors (with hyper-threading) and 192 GB RAM so that we can run our multi-threading system. We split the input into several parts according to the number of lines and start multiple processes to translate simultaneously.
Then we merge each part of translations to one file in the original order.

\subsection{Other Optimizations}
In addition to the methods above, we also tried to find the optimal settings for our system. \\
\textbf{Greedy Search} \quad
In the practice of knowledge distillation, we find that our systems are insensitive to the beam size. It means that the translation quality is good enough even we use greedy search in all submissions. \\
\textbf{Better decoding configurations} \quad
As mentioned earlier, our GPU versions use a large batch size, but the number on the CPU is much smaller. We use a fixed batch size (number of sentences) of 512 on the GPU and 64 on the CPU. We also set the number of processes on the CPU as 24 and use 2 MKL threads for each process. The maximum sequence length is 120 for the source and 200 for the target. \\
\textbf{Profile-guided optimization} \quad
To further improve our systems' efficiency, we identified and optimized the performance bottlenecks in our implementation. There are many off-the-shelf tools for performance profiling such as the gprof\footnote{
\url{https://ftp.gnu.org/old-gnu/Manuals/gprof-2.9.1/html_node/gprof_toc.html}} for C++ and the nvprof\footnote{\url{http://docs.nvidia.com/cuda/profiler-users-guide/index.html}} for CUDA. We run our systems on the WMT test set for ten times and collect profile data for all functions. Figure \ref{figure:operations} shows the profiling results for different operations on GPUs before optimizing. Before optimizing, the most time-consuming functions on CPUs is pre-processing and post-processing. We gain 2x speedup on CPUs by using multi-threads for Moses (4 threads) and replacing the Python subword tool with the C++ implementation\footnote{\url{https://github.com/glample/fastBPE}}. 

For GPU-based decoding, the bottleneck is matrix multiplication and memory management. Therefore we use a memory pool to control allocation/deallocation, which dynamically allocates blocks during decoding and releases them after the translation finished. Compared with the on-the-fly mode, this strategy significantly improves the efficiency of our systems by up to 3x speedup and slightly increases the memory usage. We further remove the $log\_softmax$ in the output layer for greedy search and other data transfers with a slight acceleration of about 10\%.  Figure \ref{figure:optimized_operations} shows the 
statistics of optimized operations. The data type conversion overhead takes about 12\% of the decoding time.

\section{Submissions and Results}
\label{sec:submissions}
We submitted five systems to this shared task, one for the CPU track and four for the GPU track, summarized as Table \ref{table:submissions}. We report file sizes, model architectures, configurations, metrics for translation, including BLEU on \textit{newstest2018} and the real translation time on a combination of test sets. The BLEU and translation time were measured by the shared-task organizers on AWS c5.metal (CPU) and g4dn.xlarge (GPU) instances.

For the GPU tracks, our systems were measured on a Tesla T4 GPU. GPU versions were compiled with CUDA 10.1, and the executable file is about 96 MiB. Our models differ in encoder and decoder layers. The base model (35-6) has 35 encoder layers and six decoder layers and achieves 44.6 BLEU on the \textit{newstest2018}. Then we see a speedup of more than one-third and a slight decrease of only 0.2 BLEU by reducing the decoder layer to 1 (35-1). We continue to reduce the number of encoder layers for more accelerations. The 18-1 system reduces the translation time by one-third with only half of the encoder layers compared to the 35-1 model. Our fastest system consists of 9 encoder layers and one decoder layer, which has one-third parameters of the 35-6 model, achieves 40 BLEU on the WMT 2019 test set, and speeds up the baseline by 3x.

For the CPU track, we used the entire machine, which has 96 virtual cores. Our CPU version is compiled with MKL static library, and the executable file is 22 MiB. We used a tiny model for the CPU with 256 hidden dimensions and kept other hyper-parameters as the 9-1 model in the GPU version. Interestingly, using half of the hidden size significantly reduces the translation quality. The main reason is that the parameters of large models cannot be reused when using smaller dimensions. This also proves that reducing the number of encoder and decoder layers is a more effective compression method. The CPU system achieves 37.2 BLEU on the \textit{newstest2018} and is 1.2x faster than the fastest GPU system.

We made fewer efforts to reduce the model size and memory footprint. Our systems use a global memory pool, and we sort the input sentences in descending order of length. Thus the memory consumption will reach a peak in the early stage of decoding and then decrease. Our base model contains 152 million parameters, and the file size is 291 MiB when stored in 16-bit floats. The docker image size ranges from 724 MiB to 930 MiB for our GPU systems, while the CPU version is 452 MiB. All systems running in docker are slightly slow down, and we plan to improve this in subsequent versions. 

\begin{table}
\begin{center}

\begin{tabular}{lrrr}
\toprule
Model & MiB & Time & BLEU \\
\midrule
Student-35-6 & 305 & 3166.4 & 44.6 \\
Student-35-1 & 264 & 2023.3 & 44.3 \\
Student-18-1 & 156 & 1355.0 & 43.4 \\
Student-9-1 & 99 & 977.6 & 42.9 \\
\midrule
Student-9-1-tiny$^{\dagger}$ & 67 & 810.9 & 37.2 \\
\bottomrule
\end{tabular}

\caption{Results of all submissions. $^{\dagger}$ indicates the CPU system. All student systems were running with greedy search. The time was measured by the organizers on their test set and we only report the BLEU on the $newstest2018$.
}\label{table:submissions}
\end{center}
\end{table}

\section{Conclusion}
\label{sec:conclusion}
To maximize the decoding efficiency while ensuring sufficiently high translation quality, we explored different techniques, including knowledge distillation, model compression, and decoding algorithms. The deep encoder and shallow decoder networks achieve impressive performance in both translation quality and speed. We speed up the decoding by 3x with lightweight models and efficient implementations. 

For the GPU system, we plan to optimize the FP16 inference by reducing the type conversion and applying kernel fusion \cite{Wang2010KernelFA} for Transformer models. For the CPU system, we will further speed up the inference by restricting the output vocabulary to a subset of likely candidates given the source \cite{Shi2017SpeedingUN, Senellart2018OpenNMTSD} and using low precision data type \cite{Bhandare2019Efficient8Q, JunczysDowmunt2019FromRT, lin2020ijcai}.

\section*{Acknowledgements}
This work was supported in part by the National Science Foundation of China (Nos. 61876035 and 61732005) and the National Key R\&D Program of China (No.2019QY1801). The authors would like to thank anonymous reviewers for their comments.

\bibliography{acl2020}

\begin{thebibliography}{26}
\expandafter\ifx\csname natexlab\endcsname\relax\def\natexlab#1{#1}\fi

\bibitem[{Ba et~al.(2016)Ba, Kiros, and Hinton}]{Ba2016LayerN}
Jimmy Ba, Jamie~Ryan Kiros, and Geoffrey~E. Hinton. 2016.
\newblock Layer normalization.
\newblock \emph{ArXiv}, abs/1607.06450.

\bibitem[{Bhandare et~al.(2019)Bhandare, Sripathi, Karkada, Menon, Choi, Datta,
  and Saletore}]{Bhandare2019Efficient8Q}
Aishwarya Bhandare, Vamsi Sripathi, Deepthi Karkada, Vivek Menon, Sun Choi,
  Kushal Datta, and Vikram~A. Saletore. 2019.
\newblock Efficient 8-bit quantization of transformer neural machine language
  translation model.
\newblock \emph{ArXiv}, abs/1906.00532.

\bibitem[{Dehghani et~al.(2019)Dehghani, Gouws, Vinyals, Uszkoreit, and
  Kaiser}]{Dehghani2019UniversalT}
Mostafa Dehghani, Stephan Gouws, Oriol Vinyals, Jakob Uszkoreit, and Lukasz
  Kaiser. 2019.
\newblock Universal transformers.
\newblock \emph{ArXiv}, abs/1807.03819.

\bibitem[{Freitag and Al-Onaizan(2017)}]{Freitag2017BeamSS}
Markus Freitag and Yaser Al-Onaizan. 2017.
\newblock \href {https://doi.org/10.18653/v1/W17-3207} {Beam search strategies
  for neural machine translation}.
\newblock In \emph{Proceedings of the First Workshop on Neural Machine
  Translation}, pages 56--60, Vancouver. Association for Computational
  Linguistics.

\bibitem[{Hinton et~al.(2015)Hinton, Vinyals, and
  Dean}]{Hinton2015DistillingTK}
Geoffrey~E. Hinton, Oriol Vinyals, and Jeffrey Dean. 2015.
\newblock Distilling the knowledge in a neural network.
\newblock \emph{ArXiv}, abs/1503.02531.

\bibitem[{Kim and Rush(2016)}]{kim-rush-2016-sequence}
Yoon Kim and Alexander~M. Rush. 2016.
\newblock \href {https://doi.org/10.18653/v1/D16-1139} {Sequence-level
  knowledge distillation}.
\newblock In \emph{Proceedings of the 2016 Conference on Empirical Methods in
  Natural Language Processing}, pages 1317--1327, Austin, Texas. Association
  for Computational Linguistics.

\bibitem[{Kim et~al.(2019)Kim, Junczys-Dowmunt, Hassan, Fikri~Aji, Heafield,
  Grundkiewicz, and Bogoychev}]{JunczysDowmunt2019FromRT}
Young~Jin Kim, Marcin Junczys-Dowmunt, Hany Hassan, Alham Fikri~Aji, Kenneth
  Heafield, Roman Grundkiewicz, and Nikolay Bogoychev. 2019.
\newblock \href {https://doi.org/10.18653/v1/D19-5632} {From research to
  production and back: Ludicrously fast neural machine translation}.
\newblock In \emph{Proceedings of the 3rd Workshop on Neural Generation and
  Translation}, pages 280--288, Hong Kong. Association for Computational
  Linguistics.

\bibitem[{Kingma and Ba(2015)}]{Kingma2015AdamAM}
Diederik~P. Kingma and Jimmy Ba. 2015.
\newblock Adam: A method for stochastic optimization.
\newblock \emph{CoRR}, abs/1412.6980.

\bibitem[{Koehn et~al.(2007)Koehn, Hoang, Birch, Callison-Burch, Federico,
  Bertoldi, Cowan, Shen, Moran, Zens, Dyer, Bojar, Constantin, and
  Herbst}]{Koehn2007MosesOS}
Philipp Koehn, Hieu Hoang, Alexandra Birch, Chris Callison-Burch, Marcello
  Federico, Nicola Bertoldi, Brooke Cowan, Wade Shen, Christine Moran, Richard
  Zens, Chris Dyer, Ond{\v{r}}ej Bojar, Alexandra Constantin, and Evan Herbst.
  2007.
\newblock \href {https://www.aclweb.org/anthology/P07-2045} {{M}oses: Open
  source toolkit for statistical machine translation}.
\newblock In \emph{Proceedings of the 45th Annual Meeting of the Association
  for Computational Linguistics Companion Volume Proceedings of the Demo and
  Poster Sessions}, pages 177--180, Prague, Czech Republic. Association for
  Computational Linguistics.

\bibitem[{Li et~al.(2019)Li, Li, Xu, Lin, Liu, Liu, Wang, Zhang, Xu, Wang,
  Feng, Chen, Liu, Li, Wang, Xiao, and Zhu}]{li-etal-2019-niutrans}
Bei Li, Yinqiao Li, Chen Xu, Ye~Lin, Jiqiang Liu, Hui Liu, Ziyang Wang, Yuhao
  Zhang, Nuo Xu, Zeyang Wang, Kai Feng, Hexuan Chen, Tengbo Liu, Yanyang Li,
  Qiang Wang, Tong Xiao, and Jingbo Zhu. 2019.
\newblock \href {https://doi.org/10.18653/v1/W19-5325} {The {N}iu{T}rans
  machine translation systems for {WMT}19}.
\newblock In \emph{Proceedings of the Fourth Conference on Machine Translation
  (Volume 2: Shared Task Papers, Day 1)}, pages 257--266, Florence, Italy.
  Association for Computational Linguistics.

\bibitem[{Li et~al.(2020)Li, Wang, Xiao, Liu, and Zhu}]{Li2020NeuralMT}
Yanyang Li, Qiang Wang, Tong Xiao, T~Liu, and Jingbo Zhu. 2020.
\newblock Neural machine translation with joint representation.
\newblock \emph{ArXiv}, abs/2002.06546.

\bibitem[{Lin et~al.(2020)Lin, Li, Liu, Xiao, Liu, and Zhu}]{lin2020ijcai}
Ye~Lin, Yanyang Li, Tengbo Liu, Tong Xiao, Tongran Liu, and Jingbo Zhu. 2020.
\newblock Towards fully 8-bit integer inference for the transformer model.
\newblock In \emph{Proceedings of the Twenty-Ninth International Joint
  Conference on Artificial Intelligence}.

\bibitem[{Ott et~al.(2019)Ott, Edunov, Baevski, Fan, Gross, Ng, Grangier, and
  Auli}]{Ott2019fairseqAF}
Myle Ott, Sergey Edunov, Alexei Baevski, Angela Fan, Sam Gross, Nathan Ng,
  David Grangier, and Michael Auli. 2019.
\newblock \href {https://doi.org/10.18653/v1/N19-4009} {fairseq: A fast,
  extensible toolkit for sequence modeling}.
\newblock In \emph{Proceedings of the 2019 Conference of the North {A}merican
  Chapter of the Association for Computational Linguistics (Demonstrations)},
  pages 48--53, Minneapolis, Minnesota. Association for Computational
  Linguistics.

\bibitem[{Post(2018)}]{post-2018-call}
Matt Post. 2018.
\newblock \href {https://doi.org/10.18653/v1/W18-6319} {A call for clarity in
  reporting {BLEU} scores}.
\newblock In \emph{Proceedings of the Third Conference on Machine Translation:
  Research Papers}, pages 186--191, Brussels, Belgium. Association for
  Computational Linguistics.

\bibitem[{Senellart et~al.(2018)Senellart, Zhang, Wang, Klein, Ramatchandirin,
  Crego, and Rush}]{Senellart2018OpenNMTSD}
Jean Senellart, Dakun Zhang, Bo~Wang, Guillaume Klein, Jean-Pierre
  Ramatchandirin, Josep Crego, and Alexander Rush. 2018.
\newblock \href {https://doi.org/10.18653/v1/W18-2715} {{O}pen{NMT} system
  description for {WNMT} 2018: 800 words/sec on a single-core {CPU}}.
\newblock In \emph{Proceedings of the 2nd Workshop on Neural Machine
  Translation and Generation}, pages 122--128, Melbourne, Australia.
  Association for Computational Linguistics.

\bibitem[{Sennrich et~al.(2016)Sennrich, Haddow, and
  Birch}]{Sennrich2016NeuralMT}
Rico Sennrich, Barry Haddow, and Alexandra Birch. 2016.
\newblock \href {https://doi.org/10.18653/v1/P16-1162} {Neural machine
  translation of rare words with subword units}.
\newblock In \emph{Proceedings of the 54th Annual Meeting of the Association
  for Computational Linguistics (Volume 1: Long Papers)}, pages 1715--1725,
  Berlin, Germany. Association for Computational Linguistics.

\bibitem[{Shaw et~al.(2018)Shaw, Uszkoreit, and
  Vaswani}]{Shaw2018SelfAttentionWR}
Peter Shaw, Jakob Uszkoreit, and Ashish Vaswani. 2018.
\newblock \href {https://doi.org/10.18653/v1/N18-2074} {Self-attention with
  relative position representations}.
\newblock In \emph{Proceedings of the 2018 Conference of the North {A}merican
  Chapter of the Association for Computational Linguistics: Human Language
  Technologies, Volume 2 (Short Papers)}, pages 464--468, New Orleans,
  Louisiana. Association for Computational Linguistics.

\bibitem[{Shi and Knight(2017)}]{Shi2017SpeedingUN}
Xing Shi and Kevin Knight. 2017.
\newblock \href {https://doi.org/10.18653/v1/P17-2091} {Speeding up neural
  machine translation decoding by shrinking run-time vocabulary}.
\newblock In \emph{Proceedings of the 55th Annual Meeting of the Association
  for Computational Linguistics (Volume 2: Short Papers)}, pages 574--579,
  Vancouver, Canada. Association for Computational Linguistics.

\bibitem[{So et~al.(2019)So, Liang, and Le}]{So2019TheET}
David~R. So, Chen Liang, and Quoc~V. Le. 2019.
\newblock The evolved transformer.
\newblock \emph{ArXiv}, abs/1901.11117.

\bibitem[{Strubell et~al.(2019)Strubell, Ganesh, and
  McCallum}]{Strubell2019EnergyAP}
Emma Strubell, Ananya Ganesh, and Andrew McCallum. 2019.
\newblock \href {https://doi.org/10.18653/v1/P19-1355} {Energy and policy
  considerations for deep learning in {NLP}}.
\newblock In \emph{Proceedings of the 57th Annual Meeting of the Association
  for Computational Linguistics}, pages 3645--3650, Florence, Italy.
  Association for Computational Linguistics.

\bibitem[{Vaswani et~al.(2017)Vaswani, Shazeer, Parmar, Uszkoreit, Jones,
  Gomez, Kaiser, and Polosukhin}]{Vaswani2017AttentionIA}
Ashish Vaswani, Noam Shazeer, Niki Parmar, Jakob Uszkoreit, Llion Jones,
  Aidan~N. Gomez, Lukasz Kaiser, and Illia Polosukhin. 2017.
\newblock \href {http://arxiv.org/abs/1706.03762} {Attention is all you need}.
\newblock \emph{CoRR}, abs/1706.03762.

\bibitem[{{Wang} et~al.(2010){Wang}, {Lin}, and {Yi}}]{Wang2010KernelFA}
G.~{Wang}, Y.~{Lin}, and W.~{Yi}. 2010.
\newblock Kernel fusion: An effective method for better power efficiency on
  multithreaded gpu.
\newblock In \emph{2010 IEEE/ACM Int'l Conference on Green Computing and
  Communications Int'l Conference on Cyber, Physical and Social Computing},
  pages 344--350.

\bibitem[{Wang et~al.(2019)Wang, Li, Xiao, Zhu, Li, Wong, and
  Chao}]{wang-etal-2019-learning}
Qiang Wang, Bei Li, Tong Xiao, Jingbo Zhu, Changliang Li, Derek~F. Wong, and
  Lidia~S. Chao. 2019.
\newblock \href {https://doi.org/10.18653/v1/P19-1176} {Learning deep
  transformer models for machine translation}.
\newblock In \emph{Proceedings of the 57th Annual Meeting of the Association
  for Computational Linguistics}, pages 1810--1822, Florence, Italy.
  Association for Computational Linguistics.

\bibitem[{Wang et~al.(2018)Wang, Li, Xiao, Li, Li, and
  Zhu}]{Wang2018MultilayerRF}
Qiang Wang, Fuxue Li, Tong Xiao, Yanyang Li, Yinqiao Li, and Jingbo Zhu. 2018.
\newblock \href {https://www.aclweb.org/anthology/C18-1255} {Multi-layer
  representation fusion for neural machine translation}.
\newblock In \emph{Proceedings of the 27th International Conference on
  Computational Linguistics}, pages 3015--3026, Santa Fe, New Mexico, USA.
  Association for Computational Linguistics.

\bibitem[{Wu et~al.(2019)Wu, Fan, Baevski, Dauphin, and Auli}]{Wu2019PayLA}
Felix Wu, Angela Fan, Alexei Baevski, Yann Dauphin, and Michael Auli. 2019.
\newblock Pay less attention with lightweight and dynamic convolutions.
\newblock \emph{ArXiv}, abs/1901.10430.

\bibitem[{Zhang et~al.(2019)Zhang, Titov, and Sennrich}]{Zhang2019ImprovingDT}
Biao Zhang, Ivan Titov, and Rico Sennrich. 2019.
\newblock \href {https://doi.org/10.18653/v1/D19-1083} {Improving deep
  transformer with depth-scaled initialization and merged attention}.
\newblock In \emph{Proceedings of the 2019 Conference on Empirical Methods in
  Natural Language Processing and the 9th International Joint Conference on
  Natural Language Processing (EMNLP-IJCNLP)}, pages 898--909, Hong Kong,
  China. Association for Computational Linguistics.

\end{thebibliography}
\bibliographystyle{acl_natbib}

\end{document}